\documentclass{article}
\usepackage{colt08e,times} 
\usepackage{amsmath}
\usepackage{graphicx}
\usepackage{amssymb}
\usepackage{epstopdf}
\newcommand{\eps}{\varepsilon}

\newcommand{\li}[1]{\lceil {#1} \rceil}

\newcommand{\lv}[1]{\langle#1\rangle}

\newcommand{\argmax}{\mathop{\mathrm{argmax}}\nolimits}

\begin{document}
\title{Statistical Learning of Arbitrary Computable Classifiers}
\author{David Soloveichik\thanks{I thank Erik Winfree and Matthew Cook for discussions and invaluable support.
}
\\ California Institute of Technology
\\ MC 136-93
\\ Pasadena, CA 91125
\\ {\tt dsolov@caltech.edu}
}

\maketitle

\begin{abstract}
Statistical learning theory chiefly studies restricted hypothesis classes, 
particularly those with finite Vapnik-Chervonenkis (VC) dimension.
The fundamental quantity of interest is the sample complexity: the number of samples required to learn to a specified level of accuracy.
Here we consider learning over the set of all computable labeling functions.
Since the VC-dimension is infinite
and a priori (uniform) bounds on the number of samples are impossible,
we let the learning algorithm decide when it has seen sufficient samples to have learned.
We first show that learning in this setting is indeed possible, and develop a learning algorithm.
We then show, however, that bounding sample complexity independently of the distribution is impossible.
Notably, this impossibility is entirely due to the requirement that the learning algorithm be computable, 
and not due to the statistical nature of the problem.
\end{abstract}

\section{Introduction}

Suppose we are trying to learn a difficult classification problem:
for example determining whether the given image contains a human face, 
or whether the MRI image shows a malignant tumor, etc.
We may first try to train a simple model such as a small neural network.
If that fails, we may move on to other, potentially more complex, methods of classification such as support vector machines with different kernels, 
techniques to apply certain transformations to the data first, etc.
Conventional statistical learning theory attempts to bound the number of samples needed to learn to a specified level of accuracy for each of the above models (e.g.\ neural networks, support vector machines). 
Specifically, it is enough to bound the VC-dimension of the learning model to determine the number of samples to use~\cite{vapnik1971ucr,blumer1989lav}.
However, if we allow ourselves to change the model, then the VC-dimension of the overall learning algorithm is not finite,
and much of statistical learning theory does not directly apply.

Accepting that much of the time the complexity of the model cannot be a priori bounded, 
Structural Risk Minimization~\cite{vapnik1998slt} explicitly considers a hierarchy of increasingly complex models.
An alternative approach, and one we follow in this paper, is simply to consider a single learning model that includes all possible classification methods.

We consider the unrestricted learning model consisting of all computable classifiers.
Since the VC-dimension is clearly infinite, 
there are no uniform bounds (independent of the distribution and the target concept) on the number of samples needed to learn accurately~\cite{blumer1989lav}.
Yet we still want to guarantee a desired level of accuracy.
Rather than deciding on the number of samples a priori, 
it is natural to allow the learning algorithm to decide when it has seen sufficiently many labeled samples based on the training samples seen up to now and their labels.
Since the above learning model includes any practical classification scheme, we term it universal (PAC-) learning.

We first show that there is a computable learning algorithm in our universal setting.
Then, in order to obtain bounds on the number of training samples that would be needed,
we consider measuring sample complexity of the learning algorithm as a function of the unknown correct labeling function (i.e.\ target concept).
Although the correct labeling is unknown, this sample complexity measure could be used to compare learning algorithms speculatively: ``if the target labeling were such and such, learning algorithm $A$ requires fewer samples than learning algorithm $B$".
By asking what is the largest sample size needed assuming the target labeling function is in a certain class, we could compare the sample complexity of the universal learner to a learner over the restricted class (e.g.\ with finite VC-dimension).

However, we prove that it is impossible to bound the sample complexity of any \emph{computable} universal learning algorithm, even as a function of the target concept.
Depending on the distribution, any such bound will be exceeded with arbitrarily high probability.
The impossibility of a distribution-independent bound is entirely due to the computability requirement.
Indeed we show there is an uncomputable learning procedure for which we bound the number of samples queried as a function of the unknown target concept, independently of the distribution.

Our results imply that computable learning algorithms in the universal setting
must ``waste samples" in the sense of requiring more samples than is necessary for statistical reasons alone.

\section{Relation to Previous Work}

There is comparatively little work in statistical learning theory on learning arbitrary computable classifiers compared to the volume of research on learning in more restricted settings.
Computational learning theory (aka PAC-learning) requires learning algorithms to be efficient in the sense of running in polynomial time of certain parameters~\cite{valiant1984,kearns1994icl}.
That work generally restricts learning to very limited concept/hypothesis spaces such as perceptrons, DNF expressions, limited-weight neural networks, etc.
The purely statistical learning theory paradigm ignores issues of computability~\cite{vapnik1971ucr,vapnik1998slt}.
Work on learning arbitrary computable functions is mostly in the ``learning in the limit" paradigm~\cite{gold1967lil,angluin1988ils},
in which the goal of learning is to eventually converge to the perfectly correct hypothesis
as opposed to approximating it with an approximately correct hypothesis.

The idea of allowing the learner to ask for a varying number of training samples based on the ones previously seen was studied before in statistical learning theory~\cite{linial1988rla,benedek1994nl}.
Linial et al~\cite{linial1988rla} called this model ``dynamic sampling" and showed that dynamic sampling allows learning with a hypothesis space of infinite VC-dimension if all hypotheses can be enumerated.
This is essentially Theorem~\ref{thm:uncomputablealgorithm} of our paper.
However, the hypothesis space of all computable functions cannot be enumerated by any algorithm, 
and thus these results do not directly imply the existence of a learning algorithm in our setting.

Our proof technique for establishing positive results (Theorem~\ref{thm:learningalgorithmexists}) is parallel evaluation of all hypotheses, 
and is based on Levin's universal search~\cite{levin1973uss}.
In learning theory, Levin's universal search was previously used by Goldreich and Ron~\cite{goldreich1997ula} to evaluate all learning algorithms in parallel and obtain an algorithm with asymptotically optimal computation time.

The main negative result of this paper is showing the absence of distribution independent bounds on sample complexity for computable universal learning algorithms (Theorem~\ref{thm:nobound}).
Recently Ryabko~\cite{ryabko2005cpr} considered learning arbitrary computable classifiers,
albeit in a setting where the number of samples for the learning algorithm is externally chosen.
He demonstrated a computational difficulty in determining the number of samples needed:
it grows faster than any computable function of the length of the target concept.
In contrast, we prove that distribution-independent bounds do not exist altogether for computable learning algorithms in our setting.

\section{Definitions}
The \emph{sample space} $X$ is the universe of possible points over which learning occurs.
Here we will largely suppose the sample space $X$ is the set of all finite binary strings $\{0,1\}^*$.
A \emph{concept space} $C$ and \emph{hypothesis space} $H$ are sets of boolean-valued functions over $X$,
which are said to \emph{label} points $x \in X$ as $0/1$.
The concept space $C$ is the set of all possible labeling functions that our learning algorithm may be asked to learn from.
In each learning scenario, there is some unknown \emph{target concept} $c \in C$ that represents the desired way of labeling points.
There is also an unknown \emph{sample distribution} $D$ over $X$.
The learning algorithm chooses a \emph{hypothesis} $h \in H$ based on iid samples drawn from $D$ and labeled according to the target concept $c$.
Since we cannot hope to distinguish between a hypothesis that is always correct and one that is correct most of the time,
we adopt the ``probably approximately correct"~\cite{valiant1984} goal 
of producing with high probability ($1-\delta$) a hypothesis $h$ such that the probability over $x \sim D$ that $h(x) \neq c(x)$ is small ($\eps$).

Here we will mostly consider the concept space $C$ to be the set of all total recursive functions $X \rightarrow \{0,1\}$.
We say that this is a universal learning setting because $C$ includes any practical classification scheme. 
We will mostly consider the hypothesis space to be the set of all partial recursive functions $X \rightarrow \{0,1,\bot\}$, where $\bot$ indicates failure to halt.
From PAC learning it is known that sometimes it helps to use different concept and hypothesis classes, if one desires the learning algorithm to be efficient~\cite{pitt1988cll}.
In a related way, allowing our algorithm to output a partial recursive function that may not halt on all inputs seems to permit learning (e.g.~Theorem~\ref{thm:learningalgorithmexists}).
Abusing notation, $c \in C$ or $h \in H$  will refer to either the function or to a representation of that function as a program.
Similarly $C$ and $H$ will refer to the sets of functions or to the sets of representations of the corresponding functions.
We assume all programs are written in some fixed alphabet and are interpreted by some fixed universal Turing machine.
If $h$ is a partial recursive function and $h(x) = \bot$ then by convention $h(x) \neq h'(x)$ for any partial recursive function $h'$ (even if $h'(x) = \bot$ also).

We can now define what we mean by a learning algorithm:

\begin{definition}
Algorithm $A$ is a \emph{learning algorithm} over sample space $X$, concept space $C$, and hypothesis space $H$ if:
\begin{itemize}
\item (syntactic requirements) $A$ takes two inputs $\delta \in (0,1)$ and $\eps \in (0,1/2)$, queries an oracle for pairs in $X \times \{0,1\}$, and if $A$ halts it outputs a hypothesis $h \in H$.
\item (semantic requirements) For any $\delta, \eps$,
for any concept $c \in C$, and distribution $D$ over $X$,
if the oracle returns pairs $(x, c(x))$ for $x$ drawn iid from $D$,
then $A$ always halts, 
and with probability at least $1-\delta$ 
outputs a hypothesis $h$ such that $\Pr_{x \sim D}[h(x) \neq c(x)] < \eps$.
\end{itemize}
\end{definition}

The always halting requirement seems a nice property of the learning algorithm
and indeed the learning algorithm we develop (Theorem~\ref{thm:learningalgorithmexists}) will halt for any concept and sequence of samples.
However, relaxing this requirement to allow a non-zero probability that the learning algorithm queries the oracle for infinitely many samples does not change our negative results (Theorem~\ref{thm:nobound}),
as long as a finite number of oracle calls implies halting.

The fundamental notion in statistical learning theory is that of sample complexity.
Since the VC-dimension of our hypothesis space is infinite, 
there is no \emph{uniform bound} $m(\delta,\eps)$ on the number of samples needed to learn to the $\delta, \eps$ level of accuracy.
We will consider the question of whether for a given learning algorithm there is a \emph{distribution-independent bound} $m(c, \delta, \eps)$ on the number of samples queried from the oracle where $c \in C$ is the target hypothesis.
In other words the bound is allowed to depend on the target concept $c$ but not on the sample distribution $D$.
Such a bound may be satisfied with certainty, or satisfied with high probability over the learning samples.

\section{Results}

We first show that there is a computable learning algorithm in our setting.

\begin{theorem}   \label{thm:learningalgorithmexists}
There is a learning algorithm over sample space $X$ of all finite binary strings, hypothesis space $H$ of all partial recursive functions, and concept space $C$ of all total recursive functions.
\end{theorem}

In order to prove this theorem we need the following lemma.
Results equivalent to this lemma can be found in~\cite{linial1988rla}.

\begin{lemma} \label{lem:m}
Let $X$ be any sample space and $D$ be any distribution over $X$.
Fix any function $c: X \rightarrow \{0,1\}$.
Suppose hypothesis space $H$ is countable,
and let $h_1, h_2, \dots$ be some ordering of $H$.
For any $\delta, \eps$, let $m(i)= \li{(2 \ln{i} + \ln(1/\delta) + \ln(\pi^2/6))/\eps}$.
Suppose $x_1, x_2, \dots$ is an infinite sequence of iid samples drawn from $D$.
Then the probability that there exists $h_i \in H$ such that $\Pr_{x \sim D}[h_i(x) \neq c(x)] > \eps$, but  $h_i$ agrees with $c$ on $x_1, x_2, \dots, x_{m(i)}$, is less than $\delta$.
\end{lemma}
\begin{proof}
The probability that a particular $h_i$ with error probability $\Pr_{x \sim D}[h_i(x) \neq c(x)] > \eps$ gets $m(i)$ i.i.d.\ instances drawn from $D$ correct is less than 
$(1-\eps)^{m(i)} \leq e^{-m(i) \eps} \leq (6/\pi^2)(\delta/i^2)$.
By the union bound, the probability that \textit{any} $h_i$ with error probability greater than $\eps$ gets $m(i)$ instances correct is less than $\sum_{i=1}^\infty (6/\pi^2)(\delta/i^2) = \delta$.
\end{proof}

\noindent\textbf{Proof of Theorem~\ref{thm:learningalgorithmexists}:}
Let $h_1, h_2, \dots$ be a recursive enumeration of $H$ (for example in lexicographic order).
For the given $\delta, \eps$, let $m(i)$ be defined as in Lemma~\ref{lem:m}.
The learning algorithm computes infinitely many threads $1,2,\dots$ running in parallel.
This can be done by a standard dovetailing technique. 
(For example use the following schedule:
for $k = 1$ to infinity,
for $i = 1$ to k,
perform step $k - i + 1$ of thread $i$.)
Thread $i$ sequentially checks whether $h_i(x_1) = c(x_1)$, $h_i(x_2) = c(x_2)$, $\dots$, $h_i(x_{m(i)}) = c(x_{m(i)})$, exiting if a check fails.  
If all $m(i)$ checks pass, thread $i$ terminates and outputs $h_i$.
The learning algorithm queries the oracle as necessary for new learning samples and their labeling.
The overall algorithm terminates as soon as some thread outputs an $h_i$, and outputs this hypothesis.
By Lemma~\ref{lem:m}, with probability at least $1-\delta$, 
this $h_i$ has error probability less than $\eps$.
Further, since $C \subset H$, the learning algorithm will always terminate.
\qed

\vspace{0.1in}
Note that it seems necessary to expand the hypothesis space to include all partial recursive functions
because the concept space of total recursive functions does not have a recursive enumeration 
(it is uncomputable whether a given program is total recursive or not).

We will see in Theorem~\ref{thm:nobound}  that there is no bound $m(c,\delta,\eps)$ on the number of samples queried by any computable learning algorithm in our setting.
Let us obtain some intuition for why that is true for the above learning algorithm.
Then we will contrast this to the case of an uncomputable learning algorithm.

In essence, we can make the above learning algorithm query for more samples than is necessary for statistical reasons alone.
Intuitively, suppose that an $h_{i^*}$ coming early in the ordering is always correct 
but takes a very long time to compute.
The learning algorithm cannot wait for this $h_{i^*}$ to finish, 
because it does not know that any particular $h_i$ will ever halt.
At some point it has to start testing $h_i$'s that come later in the ordering and that have larger $m(i)$'s.
Testing these requires more learning samples than $m(i^*)$.

If we can know which $h_i$'s are safe to skip over since they don't halt, and for which $h_i$'s we should wait,
then the above problem is solved.
Indeed, the following theorem shows that there is no statistical reason why a distribution-independent bound $m(c,\delta,\eps)$ is impossible.
The theorem presents a well defined method of learning (albeit an uncomputable one) for which there exists such a bound, and this bound is satisfied with certainty.
Below, the halting oracle gives $0/1$ answers to questions of the form $(h,x)$ where $h \in H, x \in X$ such that a $1$ answer indicates that $h(x)$ halts and a $0$ answer indicates it does not;
the answers are clearly uncomputable.

\begin{theorem}   \label{thm:uncomputablealgorithm}
If a learning algorithm is allowed to query the halting oracle,
then there is a learning algorithm over sample space $X$ of all finite binary strings, hypothesis space $H$ of all partial recursive functions, and concept space $C$ of all total recursive functions,
and a function $m: C \times (0,1) \times (0,1/2) \rightarrow \mathbb{N}$,
such that 
for any approximation parameters $\delta, \eps$,
any target concept $c \in C$,
and any distribution $D$ over $X$,
the learning algorithm uses at most $m(c,\delta,\eps)$ training samples.
\end{theorem}
\begin{proof}
Rather than dovetailing as is done for the computable learning algorithm (Theorem~\ref{thm:learningalgorithmexists}), 
we can sequentially test every $h_i$ on samples $x_1$, $\dots$, $x_{m(i)}$ because we can determine whether $h_i$ halts on a given input.
Since $c = h_{i^*}$ for some $h_{i^*} \in H$, the hypothesis $h_i$ we output will always satisfy $i < i^*$,
and therefore we will require at most $m(i^*) = \li{(2 \ln(i^*) + \ln(1/\delta) + \ln(\pi^2/6))/\eps}$ samples.
\end{proof}

We now show that for any \emph{computable} learning algorithm,
and any possible sample bound $m(c,\delta,\eps)$,
there is a target concept $c$ and a sample distribution such that this sample bound is violated with high probability.
The probability of violation can be made arbitrarily close to $1- 2(\delta+(1-\delta)\eps)$ (which approaches $1$ as $\delta, \eps \rightarrow 0$).
In fact this theorem is stronger: it shows that given a learning algorithm, without varying the target concept, but just by varying the distribution it is possible to make the algorithm ask for arbitrarily many learning samples with high probability.

\begin{theorem}   \label{thm:nobound}
For any learning algorithm over sample space $X$ of all finite binary strings, hypothesis space $H$ of all partial recursive functions, and concept space $C$ of all total recursive functions,
there is a target concept $c \in C$,
such that 
for any approximation parameters $\delta, \eps$,
for any $\rho < 1- 2(\delta+(1-\delta)\eps)$,
and for any sample bound $m \in \mathbb{N}$
there is a distribution $D$ over $X$,
such that the learning algorithm uses more than $m$ training samples with probability at least $\rho$.
\end{theorem}

The key difference between a computable and an uncomputable learning algorithm, 
is that a concept can simulate a computable one.
By simulating the learning algorithm, a concept can choose to behave in way that is bad for the learning algorithm's sample complexity.

To prove the above theorem, we will first need the following lemma.
The lemma essentially shows a situation such that any learning algorithm according to our definition must query for more than $m$ learning samples with high probability when the target concept is chosen adversarily.
The lemma is true even without requiring the learning algorithm to be computable.
Note that the lemma does not directly imply the theorem above, even in its weaker form,
because in order to increase the number of learning samples that are likely queried by the learning algorithm, 
we have to change the target concept.
Since $m(c,\delta,\eps)$ is a function of $c$, 
there is no guarantee that the bound doesn't become larger as well.

\begin{lemma}    \label{lem:probabilisticmethod}
Let $X$ be a set of $d$ points, and let $C$ be the set of all labelings of $X$.
Let $D$ be a uniform distribution over $X$.
Suppose $A$ is a learning algorithm over sample space $X$, concept and hypothesis space $C$.
For any accuracy parameters $\delta, \eps$ and any $m < d$, 
there is a concept $c \in C$ such that when the oracle draws from $D$ labeled according to $c$ 
the probability that $A$ samples more than $m$ points  is at least 
$1- \frac{2 d (\delta + (1-\delta) \eps)}{d-m}$.
\end{lemma}
\begin{proof}
We use the probabilistic method to find a particularly bad concept $c^*$.
Suppose we do not start with a fixed target concept $c$, 
but draw it uniformly from $C$.
In other words, $c$ is determined by values $\{c(x)\}_{x \in X}$ drawn uniformly from $\{0,1\}$.
Given some $x_1, \dots, x_m$, $c(x_1), \dots, c(x_m)$, and $x \not\in \{x_1,\dots,x_m\}$,
the value of $c(x)$ is a fair coin flip.
Thus if on $x_1, \dots, x_m$ labeled by $c(x_1), \dots, c(x_m)$, $A$ outputs a hypothesis without asking for more samples, 
then the hypothesis is incorrect on $x$ with probability $1/2$.
If we now let $x$ vary, the probability that the hypothesis is incorrect on $x$ is at least $(1/2)(d-m)/d$ since there are at least $d-m$ points not in $x_1, \dots, x_m$.
Now suppose for any $c$ the probability that $A$ samples more than $m$ points is at most $\rho$.
Then the unconditional probability that the hypothesis output by $A$ is incorrect on a random sample point is at least $(1-\rho)(1/2)(d-m)/d$.
This implies that there is a concept $c^* \in C$ such that the probability that the hypothesis output by $A$ is incorrect on a random sample point is at least $(1-\rho)(1/2)(d-m)/d$.

Since $A$ is a learning algorithm, 
when we use $c^*$ to label the training points, 
and use accuracy parameters $\delta, \eps$,
the probability that the hypothesis produced by $A$ has error probability greater than $\eps$ is at most $\delta$.
If we make the worst case assumption that whenever the error probability of the hypothesis is larger than $\eps$ it is exactly $1$, 
and otherwise the error probability is exactly $\eps$,
then the probability that the hypothesis output by $A$ is incorrect on a random sample point is at most 
$\delta \cdot 1 + (1-\delta) \eps$.
Thus $(1-\rho)(1/2) (d-m)/d \leq \delta + (1-\delta) \eps$,
implying that
$\rho \geq 1- \frac{2 d (\delta + (1-\delta) \eps)}{d-m}$.
\end{proof}

Now in order to prove Theorem~\ref{thm:nobound},
we essentially show that there is some fixed concept $c^*$ that behaves as the bad $c$'s in arbitrary instances of Lemma~\ref{lem:probabilisticmethod}.

\vspace{0.2in}
\noindent\textbf{Proof of Theorem~\ref{thm:nobound}:}
Consider the following program $P: \{0,1\}^* \rightarrow \{0,1\}$.
First it interprets the given string $x \in \{0,1\}^*$ as a tuple $\lv{\delta, \eps,m,d,i}$ for $\delta \in (0,1)$, $\eps \in (0,1/2)$ and $m,d,i \in \mathbb{N}$ using some fixed one-to-one encoding of such tuples as binary strings.
If $x$ cannot be decoded appropriately, or if $i > d$ then $P$ returns $0$.
Otherwise, for these $\delta, \eps, m, d$, let $\hat X \subset \{0,1\}^*$ be the set of $d$ strings 
which are interpreted as $\{\lv{\delta,\eps,m,d,1}$, $\dots$, $\lv{\delta,\eps,m,d,d}\}$,
and let $\hat D$ be a uniform distribution over $\hat X$ and $0$ elsewhere.
Let $\hat C$ be the set of all possible labelings of $\hat X$.
For each labeling $\hat c \in \hat C$,
program $P$ computes the probability $\rho_{\hat c}$ that $A$ given accuracy parameters $\delta, \eps$, queries for more than $m$ sample points if points are drawn from $\hat D$ labeled according to $\hat c$.
For each $\hat c$, this requires simulating $A$ for at most $d^m$ different sequences of sample points.
Let $\hat c^* = \argmax_{\hat c \in \hat C}\{ \rho_{\hat c} \}$, breaking ties in some fixed way.
Finally $P$ outputs $\hat c^*(x)$.

Observe that $P$ is total recursive since $A$ spends a finite time on any finite sequence of sample points.
(This is a weaker condition than the always halting requirement of our definition of a learning algorithm.)
Thus $P$ is some $c^* \in C$.
Further, for any $\delta, \eps, m, d$, on all points $\lv{\delta,\eps,m,d,i}$ for $i \leq d$,
$P$ finds the same $\hat c^*$,
and thus on these points $c^*$ acts like this $\hat c^*$.
By Lemma~\ref{lem:probabilisticmethod}, if $m < d$ then this $\hat c^*$ has the property that $\rho_{\hat c^*} \geq 1- \frac{2 d (\delta + (1-\delta) \eps)}{d-m}$.
Therefore, if $A$ is given accuracy parameters $\delta, \eps$, 
the target concept is $c^*$,
and the distribution $D$ is uniform over $\{\lv{\delta,\eps,m,d,1}, \dots, \lv{\delta,\eps,m,d,d} \}$ for some $d \in \mathbb{N}$ such that $m < d$,
then the probability that $A$ requests more than $m$ samples is at least $1- \frac{2 d (\delta + (1-\delta) \eps)}{d-m}$.
Since we can choose $D$ such that $d$ is large enough,
we obtain the desired result.
\qed

\section{Conclusion}

We have shown that learning arbitrary computable classifiers is possible in the statistical learning paradigm.
However for any computable learning algorithm, 
the number of samples required to learn to a desired level of accuracy may become arbitrarily large depending on the sample distribution.
This is in contrast to uncomputable learning methods in the same universal setting whose sample complexity can be bounded independently of the distribution.

Our results mean that there is a big price in terms of sample complexity to be paid for the combination of universality and computability of the learner.
Specifically, by tweaking the distribution we can make a computable universal learner arbitrarily worse than a restricted learning algorithm on a finite VC-dimensional hypothesis space, or even an uncomputable universal learner.

While we have presented a single computable learning algorithm in our universal setting, 
one would like to develop a measure that would allow different learning algorithms to be compared to each other in terms of sample complexity.
We have seen that sample complexity $m(c,\delta,\eps)$ is not such a measure; 
is there a viable alternative?

Finally, we have ignored computation time in our analysis.
As such, our learning algorithm is not likely to have practical significance.
Integrating running time into the theory presented would be a critical extension.

\end{document}